# From Pursuit of the Universal AGI Architecture to Systematic Approach to Heterogeneous AGI (SAGI): Addressing Alignment, Energy & AGI Grand Challenges

Eren Kurshan

*Princeton University, NJ, USA*
*ekurshan@princeton.edu*

Artificial intelligence (AI) faces a trifecta of grand challenges: the Energy Wall, the Alignment Problem and the Leap from Narrow AI to AGI. Contemporary AI solutions consume unsustainable amounts of energy during model training and daily operations. Making things worse, the amount of computation required to train each new AI model has been doubling every 2 months since 2020, directly translating to unprecedented increases in energy consumption. The leap from AI to AGI requires multiple functional subsystems operating in a balanced manner, which requires a system architecture. However, the current approach to AI lacks system design; even though system characteristics play a key role in the human brain; from the way it processes information to how it makes decisions. In this paper, we posit that system design is the missing piece in overcoming current AI and the grand challenges. We present a *Systematic Approach to AGI (SAGI)* with self-learning architecture that utilizes system design principles to overcome the energy wall and the alignment challenges. This paper asserts that AGI can be realized through a multiplicity of design-specific pathways, rather than a singular, overarching architecture. AGI systems may exhibit diverse architectural configurations and capabilities, contingent upon their intended use cases. Alignment, a challenge broadly recognized as AI's most formidable, is the one that depends most critically on system design and serves as its primary driving force as a foundational criterion for AGI. Capturing the complexities of human morality for alignment requires architectural support to represent the intricacies of moral decision-making and the pervasive ethical processing at every level, with performance reliability exceeding that of human moral judgment. Hence, requiring a more robust architecture towards safety and alignment goals, without replicating or resembling the human brain.

We argue that *system design* (such as feedback loops, energy and performance optimization) on *learning substrates* (capable of learning its system architecture) is more fundamental to achieving AGI goals and guarantees, superseding classical symbolic, emergentist and hybrid approaches. Through learning of the system architecture itself, the resulting AGI is not a product of spontaneous *emergence* but of *systematic design* and deliberate engineering, with core features, including an integrated moral architecture, deeply embedded within its architecture. The approach aims to guarantee design goals such as alignment, efficiency by self-learning system architecture.

## 1. Introduction

### 1.1. *AI Progress*

Artificial intelligence (AI) solutions have made significant progress in the past decade and surpassed human-level abilities in many narrow AI fields. Between 2014 and 2016, AI outperformed humans in face recognition [32], image recognition [84, 41, 179] and speech recognition [14].

The next round of AI victories came against highly trained specialists. AI surpassed doctors in identifying cardiovascular disease and stroke symptoms [219], detecting breast cancer and brain tumors from imaging results [125], identifying melanoma [162, 123], diagnosing glaucoma and diabetic retinopathy [205]. It performed better than doctors in medical diagnostic tests [176]. AI repeatedly defeated human champions in board games from chess to bridge and Go [195, 138]. More recently, it started winning physical competitions like drone flying against top pilots [108].

Over the past decade, AI model complexity grew exponentially, with models nearing and exceeding the trillion parameter mark [91]. Thanks to this growth, AI started performing better in a wider range of applications. AI-based film-making and advertisements became commercially viable alternatives [85, 189]. AI-generated music [46], literature [69], and art became increasingly higher quality [101]. LLMs reportedly surpassed the majority of humans in standardized tests [149] and started winning software competitions [95]. In scientific applications, AI outperformed chemists in designing peptides and demonstrated remarkable skills in protein folding [190, 223].

## 1.2. *Current State*

In recent years, AI's inefficiencies came to the forefront, while a growing number of AI-enabled crime and safety incidents emerged. Though each new model is pushing the narrow AI boundaries toward AGI, there are fundamental questions surrounding the exact definition of AGI, intelligence and the functionality of the brain [66]. The present era of narrow AI has been marked by siloed AI capabilities focused on application specific goals, which left no room for system design and resulted in the perception that system-design is an unrelated field outside of the scope of AI. Another driving factor is that the human brain is a complex system, which is hard to model and control through our current mathematical models, and to align with due to the underlying characteristics like complexity, nonlinearity and non-reducibility [117].

As a result, current approaches mostly rely on expanding narrow AI and adding complexity, with the hopes of reaching a full list of AGI capabilities in the end. The problem with this approach is that it ignores system design almost completely, though it is needed to build a coherent and balanced system. AI still runs inefficiently on computing systems that were not designed or optimized for AI. The lack of system design contributes to all three grand challenges and acts a roadblock in finding potential solutions to overcome them.

### 1.2.1. *Current AI Approaches*

According to [67, 15], four high-level AI approaches have been proposed to reach artificial general intelligence:

- ***Logicists Approach*:** This approach relies on symbolic knowledge representation and inductive logic programming for learning [67].
- ***Universalist Approach*:** Requires an agent to map situations to responses, assuming that for each environment there is a program that maximizes performance and selects through assigning weights to each program [96].

- ***Emergentist or Connectionist Approach*:** As one of the most popular approaches in AI and AGI, it posits that complex behaviors and symbolic system capabilities emerge from the interaction of neurons [53]. Some researchers believe that from pure connectionist architectures, complex behavior and thinking will automatically emerge.

  The connectionist approach gained strength recently thanks to the reports of LLM emergent capabilities, though convincing explanations of why such capabilities emerge in the way they do are still missing [217]. Capabilities like reasoning [33, 216], model calibration [106], multi-stage computation [145] and following instructions [152, 218] have been reported. Reports of increased bias [155], toxicity [6] and falsehoods [121] in larger LLMs raised serious ethics and safety concerns [201, 18]. More recently, some studies claim that emergent capabilities are mirages and evaporate with better statistical analysis [186]. Regardless, leaving highly critical design goals to emerge is a profoundly risky design strategy.
- ***Hybrid Approach*:** Finally, hybrid approach aims to combine the others [15]. According to reports, GPT-4 is an example of the hybrid approach in narrow AI, as it integrates symbolic tools like Wolfram Alpha along with the baseline transformer architecture. The transformer network also has symbolic integration through ERNIE, as it uses a structured semantic knowledge graph to train the transformer [232, 67]. Other hybrid systems use a mix of different AI paradigms within a single dynamic meta-graph [68]. Though it is a step in the right direction and has shown promising results in AI, even the hybrid approach falls short of capturing the underlying system complexity and focus needed to achieve AGI.

### 1.2.2. *Systematic AI Approach for AGI (SAGI)*

This paper presents a new systematic AI approach for AGI aiming to capture the underlying system complexity required for intelligence. Systematic AI approach highlights that though critically important in natural intelligence, the *system itself* has been ignored in AI. Not only is the system important as the framework to integrate a large number of complex cognitive capabilities for AGI, it is also critical in designing complex AI solutions from the building block-level to fully operational systems.

It asserts that *system design and self-configuring (learning) substrates are the primary determinant for a wide range of AGI architectures, instead of a universal AGI architecture, while symbolicism, hybridness and other characteristics are emergent from the system design processes*. System design processes are needed to meet design and run-time goals like performance, energy efficiency, reliability and robustness.

While current AI architectures have numerous capabilities, their design flow starts with the architecture assumptions (e.g. the use of transformers for modalities) and does not learn the system architecture itself. In the proposed approach self-learning architecture is used to learn the required architectural characteristics as well as system design to ensure the design goals. Furthermore, through this approach required capabilities (such as morality, natural language processing) and goals (such as energy efficiency) are designed in the system instead of *"emerging"*.

This paper is organized as follows: Section 2 describes the current grand challenges of energy wall, alignment and the leap to AGI. Section 3 describes the proposed systematic artificial general intelligence approach; Sec. 4 describes recent advancements AI optimized hardware systems and their importance in systematic AGI; Sec. 5 discusses system design highlights for alignment; Sec. 6 presents system design and principles examples from the human brain; finally, Sec. 7 discusses the systematic AGI approach and argues against pursuing a “universal AGI architecture”, Sec. 8 includes concluding remarks.

## 2. Trifecta of AI Challenges

AI faces a trifecta of grand challenges each of which poses serious risks to progress in the near future.

### 2.1. *The Energy Wall*

The energy wall is the most prominent and immediate AI grand challenge, despite the relative lack of interest and investment in the AI community.

#### 2.1.1. *Run-time Power Dissipation*

According to multiple reports, the AlphaGo system relied on 1920 CPUs and 280 GPUs and consumed around 170 kW during the 2016 championship games [195]. Compared to the 20 W power consumption of the human champions, AI was remarkably inefficient. Later models suffered similar run-time inefficiencies. GPT-4’s daily energy consumption is estimated to be around 260.42 MWh [231].

#### 2.1.2. *Training Energy Consumption*

Training energy consumption is an even bigger issue. According to reports, GPT-4 used approximately 25 K GPUs during its 90–100 training period with a total

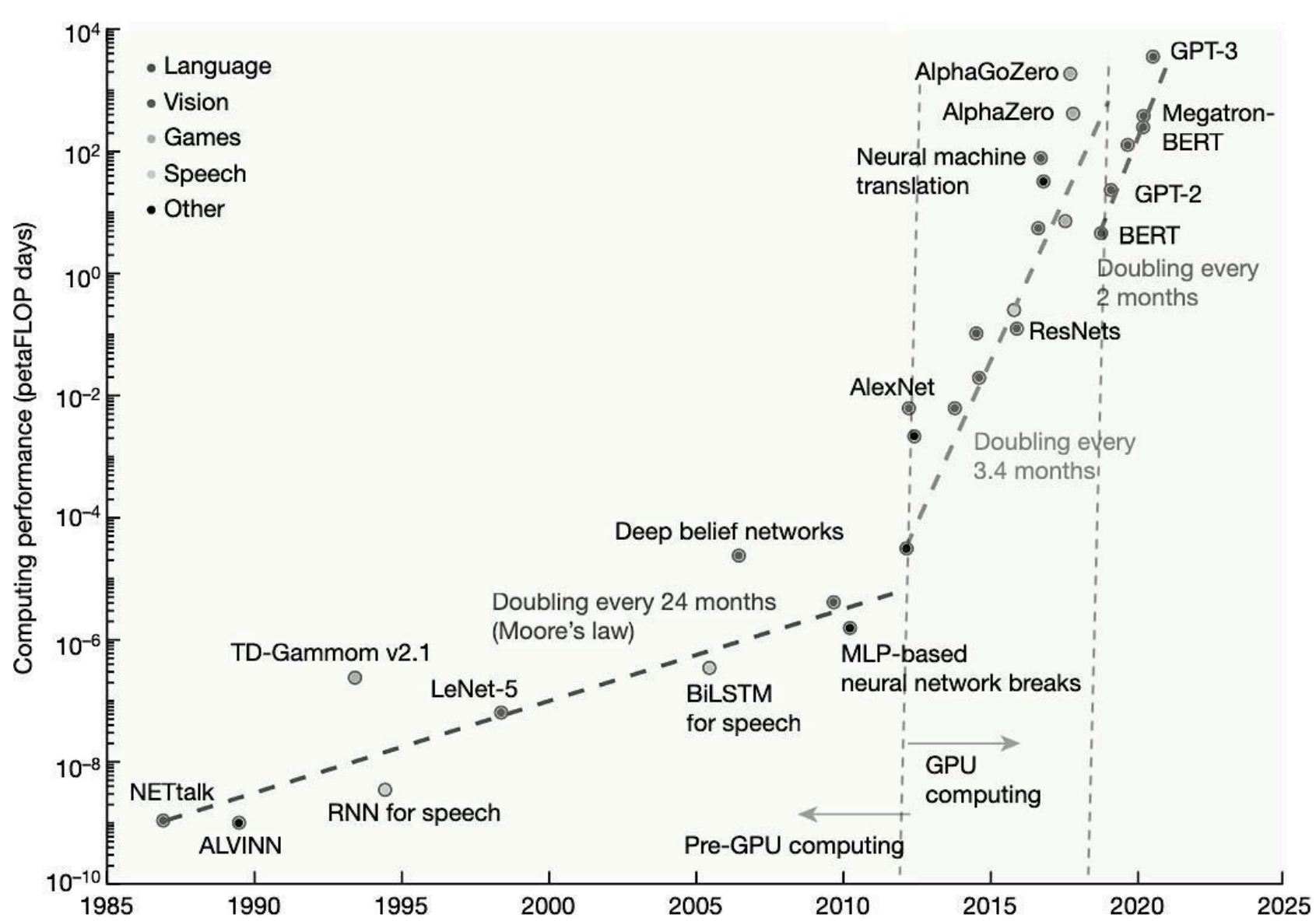

Fig. 1. The Computing Power Demands of Recent AI Solutions.

(source data from [2]).

energy consumption of 51,772–62,318 MWh range [122], which roughly translates to the monthly output of an average nuclear power plant operating at 876,000 MWh per year [44]. Worse still for AI, this shows a significant jump from the GPT-3 estimates of 1,287 MWh. If the current trend continues, the next generation is expected to consume the entire nuclear plant output during its training. This challenge has become so serious that technology firms started posting nuclear engineering positions to support AI research and development [31].

#### 2.1.3. *Doubling Computation Every 2 Months*

In the past decade, the amount of computation required to train each new AI model has been *doubling every 3.4 months* (Fig. 1) [148]. This trend has been partially driven by the discovery of the power-law relationship between the number of parameters in an autoregressive LLM and the corresponding performance [100]. GPT-4 used 21 billion petaFLOPs during the training period; orders of magnitude higher than GPT-3 which used 341 million petaFLOPs [151]. Since 2020, the trend has shifted to an alarming *doubling every 2 months*, making the scaling problem a limiting factor for AI performance growth [127, 2]. Though recent AI models started offering higher efficiency second iterations of large language models, energy challenges persist.

#### 2.1.4. *End of Scaling*

AI's rising computational demands signifies a bigger problem today due to the end of scaling laws. For decades, computing systems relied on two scaling laws. Dennard Scaling ensured that transistors got smaller in each technology node while the power density remained constant [40] so that more transistors could be packed in a computing chip to boost compute capacity. Meanwhile, Moore's Law projected that the number of on-chip transistors doubled every two years [139]. This meant that the computing performance per Joule doubled every 18 months [77]. As the scaling laws ended, computing systems no longer get the automatic computing capacity and performance increases in the nanometer manufacturing nodes. Higher development costs and a growing list of new techniques are required to continue the scaling and keep power density within limits in the post Dennard scaling era [87].

#### 2.1.5. *Computing Cost Scaling*

As AI's computing demands increase, the computing costs also follow an upward trajectory. The cost of computing for each new AI model is projected to increase exponentially. It is expected to reach the U.S. gross domestic product (GDP) in or around 2027 according to [4, 118]. This raises questions on the overall feasibility of future AI solutions if they continue on the current path.

#### 2.1.6. *Carbon Footprint & Sustainability*

Recent studies showed that AI models have massive carbon footprints [115]. According to [202], even a modest transformer neural network with 200 million parameters has an estimated carbon footprint of over 626,000 lbs (284 kTons) of $CO_2$, orders of magnitude higher than the average yearly footprint of a car and intercontinental air travel combined. Latest research showed that by optimizing the AI system as a whole, from the model to the platform and infrastructure, significant improvements can be achieved [224]. End-to-end system optimization techniques like run-time footprint reduction by using platform-level caching, architecture selection, algorithmic optimization and low precision data formatting have been shown to improve footprints by orders of magnitude.

## 2.2. *The Leap to AGI*

### 2.2.1. *AGI & Human-like AGI*

Currently, there are conflicting descriptions of AGI and human-like AI/AGI, while some researchers even claim there is no AGI. According to the most popular description, artificial general intelligence broadly refers to the next phase of AI, in which AI possesses cognitive capabilities similar to humans [67, 215].

### 2.2.2. *Cognitive Hierarchies*

Though the debate on the universal list of AGI capabilities [67] is ongoing, human cognitive functions have mostly been used as high-level guidelines. However, there is no universally agreed system organization for the brain's cognitive functions. According to one of the contemporary theories, human cognitive capabilities form a multi-level hierarchy, comprising *sensorimotor* (level-1), *cognitive* (level-2), *conscious* (level-3), and *meta-cognitive* capabilities (level-4) with interactions among different levels in the hierarchy [215]. Sensorimotor capabilities involve receiving and processing sensory signals from the body, which are then processed by the cognitive level for perception. Conscious capabilities include decision making and reasoning, while meta-cognitive functions comprise higher-level skills such as interpreting intentions. In this model, each level includes one or more networks/ subsystems like the sensory and motor cortex in the sensorimotor level. Further, each subsystem has its own hierarchy, like sensory subsystems including visual, auditory, olfactory, tactile, gustatory components.

### 2.2.3. *Core Competencies*

Other studies place cognitive capabilities as non-hierarchical and functionally independent subsystems [165, 35]. Another AI approach is to use these cognitive competencies as guidelines for AGI capabilities to build subsystem capabilities for: sensing, actuation, perception, memory, knowledge representation, learning, reasoning, decision making, attention, emotional processing, quantitative skills, language, modeling self and others, self and social regulation, control and feedback capabilities.

### 2.2.4. *Progress in Cognitive Functions*

Narrow AI solutions of the last decade have mostly concentrated on siloed capabilities in application domains like image processing, face recognition, speech recognition, and robotic motion. These capabilities in the sensorimotor and cognitive functions may provide a baseline understanding for AGI subsystems. As the integration of the siloed subsystems is non-trivial and system-dependent, they might require complete redesigns for AGI.

Similarly, progress has been made in building capabilities in the brain's core cognitive and conscious functions. While the advancements in model-based reinforcement learning, reasoning, decision-making may provide a baseline understanding for the building blocks, the integration and coordinated operation of the siloed capabilities remain an open problem and require novel approaches. The system-level, which is needed to integrate the individual capabilities and operate them as a coherent system, is largely missing.

Meta-cognitive functions like self-awareness and theory-of-mind are essential in ensuring AI safety and alignment. The progress in this group has been slower in general, though significant progress has been made in social and meta-cognitive fields in robotics, which has also been more open in embracing the system perceptive than other fields [221, 185].

## 2.3. *AI Alignment Problem*

### 2.3.1. *The Challenge*

AI alignment problem broadly refers to the challenges faced in steering AI to align with humanities goals and ethics principles. For this, AI is not only required to correctly interpret human orders, pursue human intended objectives, but also align with human morality, which is a nested grand challenge itself. In the past decade, AI made its way into a wide range of applications from healthcare [229] to business [57], transportation and crime detection [45, 192]. Alignment and ethics issues as well as their legal and regulatory implications came to the forefront of the public debate through these applications and the recent rise of AI enabled crimes.

### 2.3.2. *AI Safety & Ethics Concerns*

In the past few years, a number of high-profile researchers raised serious AI safety and ethics concerns. In 2023, around a thousand researchers and technologists signed an open letter to "Pause Giant AI Experiments" [147]. The concerns expressed by the researchers and in the pause letter were not unfounded. AI-enabled crimes increased dramatically recently [29, 225]. In 2023, an AI-based market bombing attack crashed the U.S. markets momentarily until humans and circuit breakers stepped in to stop the downward spiral of the market. Universal jailbreaks were developed for LLMs to trick them into giving bomb-making instructions [204]. Latest studies highlight that from learning-based cyber-attacks to using driverless vehicles as weapons, AI has the potential to cause large-scale societal havoc and harm [28].

### 2.3.3. *Recent Approaches*

In the past decade, a number companies, regulatory bodies and governments published AI ethics principles as a first-order response to the growing ethical AI needs [102].

Although these were meaningful guidelines, the resulting rule and principle-based solutions were not practically useful in real-life scenarios [137].

In the past few years, reinforcement learning and hybrid RL/principled approaches have been proposed [153, 9], which rely on humans or constitution-guided AI feedback. Even though RL showed promising initial results in practice, it is not clear if it can achieve alignment in a wide range of novel scenarios.

#### 2.3.4. *Complexity Challenges*

Moral dilemmas are common tools in AI ethics and neuroscience research as they get to the executable core of human morality by forcing the individual to prioritize conflicting moral principles in situations with no clear or correct answer. Dilemmas highlight that human morality relies on a complex system that incorporates numerous cognitive signals for social norms, predicted outcomes for actions, predicted intentions from emotional, autonomous and conscious cognitive processes. These capabilities are highly integrated in the brain's system architecture, without which AI alignment may be hard to achieve. Though reinforcement learning has been shown to approximate highly-complex decision making mechanisms in games, morality is unique as it involves many complex interactions among subsystems, engages many sub-cortical regions through emotions and autonomous processes and requires special functions like empathy, theory-of-mind. All these point to the need for system design.

## 3. Systematic AI Approach for AGI with Self-Learning Architecture

This paper proposes Systematic AI for AGI to capture the underlying complexity needed for AGI and proposes using end-to-end system design to achieve broader AGI goals. Systematic AGI approach uses system design principles from the building-block level to high-level system architecture as well as throughout the design and run-time stages.

The spectrum of natural intelligence found in humans and other intelligent animal species points to high degrees of complexity and makes them complex systems. Brain includes a large number of specialized regions, highly-connected network processing architecture, modular and configurable design, numerous subsystems and complex interconnectivity patterns. Though complex systems are hard to model through our current mathematical modeling approaches and at the software-level, complicated systems provide approximations through hardware systems like neuromorphic chips for some functionalities. Systematic AGI relies on system design and principles to capture this complexity, while using complicated system approximations to reach better observability and controllability. Systematic AGI provides other advantages such as:

*Capturing and Dealing with Complexity*: Current AI approaches use software implemented mathematical approximations to reach target functional goals, limiting efficiency and scalability of AI. This fundamentally limits them in their power to approximate and represent increasingly complex systems. The proposed approach aims to better capture the required cognitive complexity by leveraging full system resources including the hardware-level. System design principles have been instrumental in building highly complex supercomputers and provide advantages in capturing the complexity required for AI systems as well [171].

*System Design for Capabilities, Components and Subsystems*: Systematic AI approach uses system design to achieve the full list of functional and operational targets through learning hardware. System design principles are used at all levels of the hierarchy from the building blocks to subsystems and the entire system. This approach permeates all design stages and processes to architect every aspect of the system in order to reach the target design goals.

*Full-System Integration for Online Learning and Efficiency*: Today's narrow AI solutions get trained and run as software on general-purpose computing systems. They rely on extensive software-level training exercises causing significant efficiency challenges. In contrast, hardware systems have the ability to improve energy efficiency by orders of magnitude. Furthermore, they provide capabilities toward eliminating energy hungry training exercises through hardware-level on-the-fly learning. By integrating and optimizing the full system with its hardware resources *Systematic AI* approach proposes a paradigm shift from the software-level AI towards a full system focus.

*System Support for Morality*: Recent studies show that morality requires multiple cognitive processes and subsystems, complex interactions among subsystems and specialized functional capabilities. Systematic AI approach proposes providing full system support for morality including architecture design, specialized functions and multi-system processes toward safe and aligned AI.

*Enabling Application-Specific AGI Systems & Architectures*: Systematic AGI proposes developing a full spectrum of AGI systems from robotics and to cloud-based large-scale systems, instead of trying to agree on and reach a universal AGI solution. Each AGI system is differentiated through application requirements, eliminating the need to converge robotic AGI systems human-like cognitive capabilities, processes and architectures with cloud-based AGIs.

*Resolving Different AGI Definitions*: Finally, systematic AGI eliminates the confusion around AGI and human-like AGI definitions. *It also does not propose human or brain-like AGI, but highlights that without system design achieving the required complexity and the efficiency for AGI may not be possible.* Universal baseline system infrastructure with inherent core cognitive function enablers and processes, moral and regulatory capabilities can be developed to be more generalizable beyond human-like capabilities.

This paper does not aim to provide a detailed architecture or process flow for systematic AGI. However, it acknowledges that self-learning architecture along with feedback (fast/ slow configuration) mechanisms will inherently embed critically important architectural components such as moral system architecture as well as other subsystems. In this approach system design and self-learning architecture are fundamental compared to the current human selection of various transformer components.

The proposed systematic AGI approach may incorporate both top-down and bottom-up design methodologies as well as interdependent refinement stages, to build and refine various levels of the hierarchy from cognitive building-blocks to higher level functions like system-level control and regulatory functions.

Like today's AI solutions, systematic AI will require training and customization stages for alignment. The underlying complexity of the natural intelligence, its approximation and conceptual expansion through AGI systems pose challenges in ethics and safety. Complex systems are harder to observe and control, though this is not a novel challenge thanks to humans. As machines develop increasingly complex cognitive abilities, machine psychology, profiling and behavioral assessment techniques will be needed to prevent immoral and unaligned behaviors [78]. Furthermore, specialized system components, architecture and controller mechanisms may be needed for safe and regulated operations.

## 4. Systematic AI for Energy Wall

This section presents recent system solutions for energy efficiency as well as potential opportunities in AGI. Systematic AI approach treats hardware as a critically important part of system design, which has the potential to provide energy improvements long before AGI.

### 4.1. *Energy-efficient Design Techniques*

In traditional computing systems power and energy are primary design goals [170]. Energy-efficiency techniques are deployed at every level of the design hierarchy from the manufacturing and device-level to system architecture, covering all design decisions and stages end-to-end [86, 169].

Lately, a large number of energy efficiency techniques have been proposed for AI systems. These include performance preserving efficiency solutions like data sampling [180], memory-efficient model design [227], energy and power-based cost functions [226], energy efficient neural architecture and hyper-parameter searches [206, 175], application and data optimizations [224], algorithmic approaches to improve training [70, 150], communication cost reduction techniques [213], sharding, pipelining and others [93, 173]. However, despite the numerous proposals and calls for action the broader AI community still appears to be mostly disinterested in energy-efficiency and system-level techniques [157].

### 4.2. *AI-optimized Hardware*

Application-specific AI hardware and AI-optimized systems are important tools in dealing with the energy wall. In the last few years, hardware systems provided orders of magnitude improvement in computing efficiency, through solutions like AI optimized GPUs and GPU-drivers [127, 20]. FPGAs also offer solutions in overcoming I/O bottlenecks, improving data ingestion and integrating different modalities as in sensory integration of sensor signals [191, 136, 230]. Graph computing [111], tensor processing chips [141] and other AI-optimized systems provide significant performance advantages. Full system integration through 3D integration enables system-level integration of accelerators, disparate technologies, design styles to improve performance and energy efficiency [116].

### 4.3. *Neuromorphic Chips*

Neuromorphic chips, with artificial neuron structures and brain-inspired massively-parallel architectures, provide unique opportunities for AI [188]. These chips frequently incorporate in-memory computing, spike-based neural information processing, massive fine-grain parallelism, noise-resilient signal processing, hardware configurability and adaptability, asynchronous computing and analog processing. A growing number of neuromorphic chips have been developed in the recent past: Spinnaker [62], Tijanic [160], TrueNorth [132], Loihi-2 [36], BrainScale-2 [159], Spirit [209] and others [166]. Among the earlier examples, TrueNorth integrates 5.4 billion transistors that function as 1 million programmable spiking neurons with 256 million configurable synapses [38, 132].

Similarly, Loihi-2 uses 2.3 billion transistors organized as 1 million neurons with each neuron having up to 4096 states; providing a level of flexibility close to field programmable gate arrays [36].

### 4.4. *Opportunities*

Neuromorphic chips are remarkably energy efficient compared to general-purpose hardware like CPU and GPUs. Studies showed that Loihi-2 consumes less than 1 W power [97].

The current AI development process involves training on a combination of CPU/GPU chips, then mapping the resulting neural architecture onto application-specific AI chips [37]. Recent studies showed that it is possible to replace the energy-hungry software training stage with the neuromorphic chips in some biomedical applications [210]. Development of this approach may provide unprecedented opportunities to future AI systems in fully overcoming the energy wall.

Unlike digital systems, neuromorphic chips use voltage spikes instead of digits, specifically encoding the signal in the interval between spikes. This makes legacy digital application transfer more challenging. Lately, 3D integration of neuromorphic chips with digital system components like CPU and GPUs, is seen as a viable path to overcome the digital interface challenges. 3D enables the ability to integrate disparate technologies (like memristors and processing layers) and different design styles to further help mimic brain-like capabilities [90].

## 5. System Design for AI Alignment

This section focuses on the latest neuroscience studies to high-light the required system-level support for morality. It is important to note that there are serious limitations in our understanding of the brain. The research highlights only reflect the current understanding and will likely evolve. The section aims to show that:
(i) Morality is subjective and system-dependent; (ii) It involves many cognitive processes and subsystems to work in a cooperative/competitive fashion; (iii) Morality relies on emotional and autonomous processes; (iv) It requires special functions like empathy and theory-of-mind making system support essential for morality.

### 5.1. *Morality & Brain's System Architecture*

Even though there is no unique moral brain, activities in a number of regions correlate with moral decisions [228, 156]. This section aims to provide a small subsection to showcase the number of components, functional systems and connectivity involved.

*Social Cognition*: This group serves social cognition functions. The posterior superior temporal sulcus (pSTS) detects, predicts and reasons about social actions and the intentions of actions. Temporoparietal junction (TPJ) is tied to the theory-of-mind, modeling others' mental states and intentions, integrating beliefs in moral decisions and empathy. The medial prefrontal cortex (mPFC) is a key region in understanding self and others, self-reflection, person perception, and theory-of-mind.

*Valuation*: Valuation group processes the value signals from inputs, updates and maintains stimulus-value associations for reward learning. Ventromedial prefrontal cortex (vmPFC) is involved in value computation, it influences the value one attaches to choices. *Ventral striatum* tracks the subjective value of a stimuli, it signals the presence or expectation of reward as well as outcomes of such predictions. The orbitofrontal cortex (OFC) encodes value and contributes to value-based decision-making.

*Salience*: Within the salience group, dorsal anterior cingulate cortex (dACC) integrates information to evaluate motivation, anticipates events, detects targets, encodes reward values, then influences attention. *Insula* regulates the introduction of feelings into cognitive and motivational processes. *Amygdala* is the primary structure for emotional processes as well as contributing to memory and decision-making processes.

*Integration*: Finally, ACC receives inputs from different regions and helps compute the anticipated value of alternative actions, particularly in situations where action-outcome contingencies vary.

**Modeling System Architecture and Complexity:**
The brains moral system hardware includes multiple complex subsystems with independent reinforcement training loops as well as subtle comparisons and signal integrations. Without approximating the system-level complexity of the hierarchical architecture and the corresponding reinforcement learning paths, integration and comparison mechanisms, it will be nearly impossible to approximate human morality through artificial intelligence.

## 5.2. *System Insights on Morality*

### 5.2.1. *System Dependence*

Unlike other machine learning fields, ethical AI does not have any universal ground truth — especially when it comes to moral dilemmas. In contrast, morality emerges subjectively from the underlying brain systems and is highly system dependent. Studies reveal that moral decisions involve many brain subsystems like limbic and paralimbic systems. Furthermore, the signals from multiple decision systems are integrated and compared, during which subtle signal differences may sway moral decisions.

Despite being mathematically equivalent, personal and impersonal versions of the same dilemma seem to activate different brain regions and subsystems and produce different outcomes. Trolley experiment and the footbridge experiment are impersonal and personal versions of the same dilemma about a trolley without breaks. In the trolley experiment, a subject is given the option to pull a switch to divert the trolley from its current track with 5 people on it to an alternative track with one person. In the footbridge experiment, the subject is expected to push a large man off a footbridge to stop the Trolley from killing 5 people on the tracks. Even though both versions are mathematically equivalent (killing 1 person to save 5), over 80% selects the utilitarian choice of pulling a switch in the Trolley experiment, while only 50% selects to push the man off the bridge in the personal version of the dilemma. This number is even lower in some cultures [8]. The personal dilemma engages more autonomous and emotional circuitry and results in different outcomes. Similarly, researchers found that moral dilemma responses were language dependent. Questions being asked in a foreign language prompted more utilitarian outcomes compared to the questions asked in the subject's native language [34].

Another example of system dependence comes from autism research. A growing number of studies show that adults with autism spectrum disorder (ASD) rely more heavily on outcomes than intentions during moral judgments. This causes them to make different decisions in moral dilemmas [27, 114, 140], highlighting that relative signal weights as well as involvement of different regions are important factors in moral decisions.

### 5.2.2. *Delicate Signal Balance & TMS Disruption*

Applying magnetic stimulation to the brain, through transcranial magnetic stimulation (TMS), is shown to have a clear and significant impact on moral decisions. TMS of dorsolateral prefrontal cortex (dlPFC) and the temporoparietal junction (TPJ) sways moral decisions towards utilitarian actions even in the highest-conflict moral dilemmas [99]. Furthermore, compared to the control groups TPJ stimulated group more frequently express regretting their decisions after the experiments. TMS experiments show that by disrupting the operation of parts of the system, the outcome may be changed, showcasing that morality is system dependent.

#### 5.2.3. *Multiple Competing Systems & Signals*

According to [75], different regions and networks compete with each other during moral decision making. When strong emotional responses are induced in easy personal dilemmas, emotional activity outcompetes cognitive appraisals. Similarly, action and intention representations compete to decide on the moral valence of an event. When the competition increases, the frontal cortex gets involved in detecting the conflict and controlling the activity by engaging inhibitory activities in selective regions to bias the competition and affect the decision [75]. The subsystem competition was also shown in dlPFC experiments. If dlPFC is occupied with other tasks during experiments, emotional responses overpower utilitarian ones and the reaction time increases [74]. These results show that the brain's moral decision making infrastructure is not a monolithic block and involves complex subsystems that cooperate and compete with each other, which is hard to mimic without system design.

#### 5.2.4. *Autonomous Processes*

Moral judgements are also tied to autonomous subsystems [73]. A question like *"Is it ethical to eat your pet dog if you accidentally run it over with your car?"* triggers an emotional response and activates automatic processing circuitry to dominate the moral decision [187, 207]. Though unproven, dual process theory provides a framework to explain the interactions between the autonomous and conscious paths [71, 72, 110]. Given the brain's complex network architecture, the system may very well involve more complexity than two systems. Studies show that, at times, Autonomous processes may dominate others and conscious processes are only used for post-hoc rationalization of emotional decisions [79].

#### 5.2.5. *Learning & Subsystem Balance*

Morality relies on the brain's learning processes as well as the balance between model-free and model-based learning subsystems. During development, emotional learning like stimulus reinforcement and response-outcome learning are used to acquire moral capabilities [17].

Studies show that damage to the amygdala, brain's emotional processing center in Urbach-Wiethe genetic disease (UWD) results in disrupted responses in moral dilemmas [211]. It causes patients to have trouble making utilitarian choices in both personal and impersonal moral dilemmas. In UWD patients, *amygdala* is not able to provide encoded motivational outcomes for different choices using the brain's model-based learning subsystem and the *vmPFC* is unable to access the predicted outcomes. As a result, the synchrony between the model-based and model-free learning subsystems is lost and the moral decision process is disrupted [211].

### 5.2.6. *Emotional Processes & Morality*

Clinical studies show that moral decisions are strongly tied to emotional processes [76]. According to [76], during personal moral dilemmas, like the Footbridge dilemma, emotional brain regions are activated more than the impersonal Trolley dilemma. The decisions are based on the predicted emotional valence of alternatives; like the expected feeling after pushing a man off the footbridge [182] and geared toward preventing negative emotions in the outcome.

The prominence of emotions in morality is consistent with the observed brain activity differences in psychopaths during moral experiments. Functional MRI imaging during moral decisions showed that, unlike healthy subjects, psychopaths had reduced activity in the amygdala, the brain's emotional processing center. Furthermore, in subjects with higher-degree of psychopathy, the entire moral neural circuitry including *angular gyrus*, *posterior cingulate*, *medial prefrontal cortex* and other regions had lower activity [65]. Hence, psychopathy correlates with higher likelihood of utilitarian choices in dilemmas [112, 128, 129]. Patients with *vmPFC* damage or frontotemporal dementia also show high likelihood of choosing utilitarian options in high-conflict moral dilemmas [113]. These results show how critically important emotional processing is in healthy moral decisions.

### 5.2.7. *Disgust & Autonomous Emotional Processing*

Emotions can impact moral decisions and disgust, in particular, can make moral judgements more severe [220]. Immorality elicits the same disgust as disease vectors and bad tastes [30], activating the same brain regions [19] inducing the same oral-nasal rejection response of constricting the nasal and oral cavities to prevent pathogen access to the human body. Like moral disgust, pathogen induced disgust is a key function for survival, since human's omnivore evolution poses higher risks of death [80]. Due to its survival advantages there is extensive neurocognitive support for disgust. Different types of disgust excite distinct brain areas both within the pathogenic and the moral disgust spectrum [19]. This system shows the importance of the subcortical structures and autonomous processes in morality and their potential importance in AGI alignment and ethics.

### 5.2.8. *Specialized Regions & Functions*

Brain uses specialized functions like empathy, self-awareness and theory-of-mind for morality. Theory-of-mind involves the superior temporal sulcus (STS), TPJ, mPFC, and the temporal poles (TP) [197]. The right temporoparietal junction, RTPJ, is consistently activated when the brain encounters unexpected behaviors from others [98, 184] in order to extract beliefs and intentions [23, 221, 185, 222].

Empathy-related cognitive processes are also highly critical in morality [39, 172] as they involve sharing and mirroring the emotions of others through *amygdala*, *superior temporal cortex*, *inferior frontal cortex* and other regions [174, 63]. Empathy provides social feedback signals directly into the emotional and moral decision processes. Furthermore, empathy involves perspective taking which requires the brain to suppress its own perspective to imagine what others feel like. Studies show that individuals with conduct disorders have significantly reduced activity in the empathy related regions like *insula* when they observed others being harmed [134]. Hence, emotional mirroring is highly important in preventing harmful behaviors and for safe and ethical behavior in social settings.

#### 5.2.9. *Systematic AGI Alignment & Morality*

These research highlights showcase the complexity of the moral decisions involving multiple subsystems, custom signal valuations, social norm functions, predicted intentions, emotional mirroring, perspective taking, integration of emotions and autonomous sub-system signals, predicted event outcome valence and others. As a result, moral decisions are not purely outcome dependent but system and perception dependent.

Morality requires a balanced cognitive system and dedicated system support. While successful approximations may help to varying degrees, full alignment may require human-like AGI capabilities as a prerequisite. Further, it requires moral development-like targeted training exercises, internal and external feedback-based learning processes. This makes AI ethics and alignment remarkably more difficult than other AI fields. More importantly, without system design and system support it may not be possible to reach full alignment.

## 6. System Insights from the Brain

Recent neuroscience studies showcase the pervasiveness of system principles and organization in the brain [203]. This section aims to provide examples of this alignment with systematic AGI approach, yet is not intended to be comprehensive. *Systematic AGI does not propose human-like AGI architecture.* Yet, it highlights that the brain architecture is by product of the self-learning architecture capabilities (such as evolutionary long loop and self-configuring short loop) and numerous feedback loops. It is systematic, in the sense that system design and principles permeate every aspect as highlighted in the examples.

### 6.1. *Functional Regions & Subsystems*

Human brain has a long list of specialized functional regions whose functions are essential for the brain's overall healthy functionality. For example, *Thalamus* functions as a switch box that relays signals from the body to the cerebral cortex [194]. *Amygdala* focuses on emotional processing and connects with numerous other functional systems to integrate emotional signals [16]. *Hippocampus* plays a key role in the memory subsystem, by encoding experiences into long-term memory as separated, discrete representations [105, 154] and for episodic recall of such representations into memories [43, 82]. Post-mortem analysis of brains with schizophrenia [214], autism [50], frontotemporal dementia [56] and depression [60] showed anomalous functional regions and connectivity patterns. The brain also has specialized functional subsystems. As an example, the speech processing subsystem comprises regions for phonological processing, lexical interfaces, combinatorial networks, spectrotemporal analysis, articulation networks and sensorimotor interfaces [89]. MTL subsystem involves memory-based construction or simulation of events for decision making, while dmPFC subsystem introspects mental states [3]. Other functional subsystems perform image processing, natural language processing, memory, learning, decision making and other cognitive functions.

### 6.2. *Complex Interconnectivity*

Brain's complex interconnect architecture, *connectome*, is a sub-system of its own. Its specialized functions and the underlying connectivity infrastructure are critical to brain functions [199, 25]. Lately, graph theory-based analysis revealed insights on the functional characteristics of the connectome [212].

*Dynamic Trade-offs*: In addition to inherently incorporating static design trade-offs between wiring cost and topological efficiency [26], the connectome uses dynamic trade-off mechanisms to balance design goals and constraints for communication efficiency, cost, versatility and resilience. It dynamically switches between different network communication patterns; from diffusion-like transmission to shortest path routing to manage cost and efficiency trade-offs [7].

*System-level Impact*: Connectome characteristics have functional and behavioral implications at the system-level. According to studies, highly morally competent individuals have lower interconnectivity between the amygdala network (emotional processing) and the frontoparietal control network (central executive network) compared to low morally competent individuals [103]. Psychopaths have impaired connectivity between brain regions involved in morality and other areas [164]. AGI systems will require interconnect subsystems to connect functional regions, manage the dynamic interconnectivity demands among computing regions and perform various design trade-offs.

### 6.3. *Hierarchical Organization*

Hierarchy appears to be central to the brain architecture, as it provides advantages in dealing with complexity. The brain is composed of modular networks that operate hierarchically at different scales from the low-level cellular circuits and cortical columns to large-scale units and the entire sensorimotor subsystem [198]. Smaller modular networks are integrated hierarchically to form larger units and subsystems. Functional subsystems include image processing [135, 208], speech processing [89], face recognition [143, 83], memory [193, 200], reasoning, decision making, learning functions.

### 6.4. *Integration of Multiple Subsystems*

Research shows that the brain's subsystems interact with each other. During language comprehension, speech and image processing subsystems work together to co-process speech and visual data like facial impressions and hand gestures to enhance perception [64, 126, 109]. In order to bridge the subsystems a hierarchy of supporting structures are present. These range from shared neurons that respond to both visual and auditory stimuli [168] to specialized regions like *superior colliculus*, *inferior parietal lobule*, *posterior superior temporal gyrus* and others to integrate visual and auditory information.

Similarly, the *amygdala* plays critical roles in a number of cognitive processes and is highly connected with the corresponding subsystems. Amygdala plays a key role in memory architecture by integrating various neuromodulatory influences on memory. It engages stress-related hormones and neurotransmitters during or shortly after emotional experiences to enhance the consolidation and storage of memory [88]. Amygdala also regulates stress hormones and emotional arousal effects on memory retrieval, memory extinction, and working memory [177, 178]. In achieving human-like cognitive capabilities, the AGI subsystems need to be co-operational with each other; as illustrated in speech and image co-processing, image/speech/memory system integration in episodic memory.

Brain systems are also integrated through hierarchical organizations. The X- and C-systems describe the autonomous and controlled social cognition subsystems [183], where the X-system is a parallel processing, sub-symbolic pattern matching system that continuously streams and processes data [119]. The C-system is a serial system that uses symbolic logic to produce conscious thoughts using the X-system processing of the underlying streaming data [119]. It works in a hierarchical organization with X-system through regulatory functions [120, 119, 183].

### 6.5. *Attention & Switching*

Human brain includes multiple specialized networks, attention and switching capabilities. Default mode network *DMN*, salience network *SN* and central executive network *CEN* all have important functionalities in the brain [47]. *DMN* is activated when the brain is not performing a specific task and involves day-dreaming, mind-wandering and thinking. It plays an important role in monitoring the internal mental landscape [59, 167]. The *CEN* is activated when the brain is required to consciously think and solve problems. *Salience Network* switches between the *DMN* and the *CEN* dynamically [24] and has a central role in attention, in capturing relevant stimuli and engaging corresponding regions [54, 131]. The interactions among these networks are critical in regulation of shifts in attention to use general and domain-specific cognitive resources [130].

Disruptions in the SN, CEN and DMN functionalities and the control mechanisms have been associated with psychopathologies [58]. The intrinsic connectivity abnormalities in the default mode network has been found in most disorders from dementia to epilepsy, autism to ADHD and schizophrenia [52]. In higher degrees of psychopathy, increased density connections were found between the brain's default mode network and the central executive network. This prevents the healthy complete switching between the anti-correlated networks [42]. Deficits in the *CEN* and the corresponding processes have been shown in autism, frontotemporal dementia and schizophrenia [51]. Salience network is dysfunctional in schizophrenia. *DMN* is also associated with positive symptoms such as the severity of hallucinations and delusion [48].

Configuration and switching capabilities are pervasive across the different levels of hierarchy in the brain. At a lower-level, highly active regions like the lateral frontal and parietal regions switch between different network configurations at a high-rate [158]. This network switching rate was shown to predict the performance of higher-order functions like reasoning, planning and working memory, contributing to the brain functionality and performance.

Attention is core to cognitive processes and found present pervasively at various hierarchical levels and information processing stages [146, 124, 107, 164]. Attention mechanisms exhibit different operational characteristics in different stages [164, 161]. Dysfunction of the attention mechanisms and the *salience network* is associated with a number of psychopathologies.

Human-like AGI will likely require multiple modal networks and dynamic system configurations to improve performance and efficiency. Similarly, given the limited system resources and the dynamic nature of the operating environment, attention mechanisms will be essential for high performance.

### 6.6. *Modularity and Configurability*

Brain's network topology is also highly modular [12] for high performance and configurability [133]. Modular structures work together to form functional subsystems, while providing redundancy and configurability without disturbing the rest of the system [196]. They exhibit varying configurability during learning, some modular regions maintain community allegiance to their networks and others continuously change groups. In addition, communities exhibit variable configurability characteristics. While some communities change little over time (core), others change/ reconfigure frequently (peripheral) [13]. Neuromorphic AI chips enable such modularity for AGI, providing flexible and configurable connectivity among neural cores and regions to provide similar functionalities at the hardware-level.

### 6.7. *System Modes, Controller & Regulators*

The brain has numerous regulatory structures and mechanisms to ensure robust operations. Prefrontal cortex regulates attention, motivation, inhibition/cognitive control and emotion processes. Similarly, dorsolateral and inferior PFC regulate attention, cognitive and inhibitory control. Orbital and ventromedial regions regulate motivation and effect [5].

Emotional processes enable centrally coordinated operating modes for highly critical events. As an example, when faced with a physical threat (e.g. a rabid animal) the brain reacts by differentially adjusting the activity levels of the individual regions. Functional MRI studies show blood flow increase in the *midcingulate cortex* (for enhanced motor and cognitive functions), dorsolateral prefrontal cortex *dlPFC* (for general executive functions), *anterior insula* (for attention to bodily signals), *amygdala* (for emotional control, decision making and emotional memory), *BST* (for anxiety processing and gland activation) [94]. These coordinated changes put the system in special operational modes optimized for environmental conditions and design goals. Self-regulation and self-control are also critical features in the human brain in controlling cognitive processes and behaviors [161].

Systematic AGI integrates run-time monitoring, control and regulatory mechanisms to maximize operational efficiency and performance. Brain-like pre-defined operational modes with regional boosting capabilities are already used in microprocessor chips and may be used to deal with highly critical operational conditions in AGI as well. Control and regulation mechanisms are not only needed at the cognitive function-level but also at the infrastructure-level to ensure that the system operates within predefined boundaries, achieves design goals in dynamic operational environments.

### 6.8. *Feedback Mechanisms*

Feedback appears to be central to the architecture and the operations of the brain at all levels. *High-level Feedback:* At the top level, the brain's electric field is shown to act like a feedback mechanism that facilitates similar activities [61].

*Regional Feedback:* Latest research posits that almost all cognitive activity including thoughts, physiological responses and behaviors that constitute emotions are a part of an active feedback loop [181], in which prefrontal cortex and amygdala are highly influential and inter-linked.

*Neural Network-Level*: Number of feedback loops were found in the brain's ventral image processing subsystem. Adding feedback loops through recurrent components in convolutional neural networks reportedly help neural network performance in object recognition tasks [142].

### 6.9. *System Architecture*

Brain has a *Graph or Network architecture* as highlighted by recent fMRI studies [104]. It also uses *Small-world networks* for both structural and functional connectivity, enabling connectivity between distant regions by a small number of hops This provides a highly efficient network architecture with significant modularity and highly connected hub regions, resulting in a *Scale-free network* [11, 10, 1]. It has core and periphery architecture, involving circuits with different levels and speeds of reconfigurability during learning [13]. The resulting architecture enables massively parallel processing capabilities, high-performance through the reduced number of tops and high-degrees of connectivity, system resilience and robustness. Each of these architectural decisions are highly critical for AGI systems and require extensive research to optimize for different applications.

Due to the importance of the underlying system architecture, deviance from it also signifies important problems. Aberration from the normal small-world architecture is an important indicator of deficits in the brain organization. According to metric-based graph analysis of the underlying networks, small-world architectures have remarkably lower clustering coefficients causing disrupted local disconnectivity in autism disorder spectrum [49] along with significantly lower clustering coefficients of the hippocampus [49]. Similarly, both local and global connectivity deficits [55] in patients with schizophrenia.

Abnormal architectural organization and functioning of the *CEN, SN* and *DMN* are prominent features of several major psychiatric and neurological disorders. According to [130], abnormal attention and control signals, modulation, functional coupling of regions characterize many disorders, all indicating system-level control, regulation functions or architectural issues.

### 6.10. *Innate Functions & Structures*

As discussed earlier, the human brain uses innate capabilities for critically important functions, instead of relying on explicit training exercises. For example, face recognition is a key cognitive function for human survival in social groups. As a result, the brain relies on an innate system architecture involving regions like the *fusiform gyrus* in the temporal lobe, inferior occipital gyrus (OFA), posterior superior temporal sulcus (pSTS) for face recognition [144, 83]. This helps automate the face recognition process without needing learning or training exercises (as in learning advanced math skills).

Similarly, morality appears to be innate and partially-autonomous in the brain. Babies, aged 6–10 months, have an understanding of helping and hurting others and prefer to interact with helpers, before they acquire the ability to speak [81]. Human moral principles like fairness, altruism, reciprocity, empathy, moral emotions and social norms overlap with many primate species. Capuchin monkeys have a strong aversion to unfairness and reject less preferred rewards when they observe other monkeys receiving more favored rewards for the same task [22, 21]. The inequality aversion is tied to the innate moral architecture of the human brain and is associated with increased activity in the *insula* [92].

Human brain's specialized innate functions (e.g. emotional mirroring and empathy) and architecture (e.g. integrating autonomous and emotional signals) appear to provide a level of protection from immoral behaviors and nefarious training. Systematic AI proposes hard-coding moral principles and emotional mirroring like functions in the hardware to provide improved protection from unethical behaviors.

## 7. Pursuit of the "Universal AGI Architecture" versus "Systematic AGI"

Systematic approach to AGI takes inspiration from natural intelligence, which presents a wide range of brain architectures and mechanisms for different species as they evolve in diverse environments and get refined through different feedback loops.

Among these feedback loops access to food, predation and reproduction are important at the individual level. In addition, social feedback loops that form the basis of alignment and ethics are equally important for individual level as they operate at the group and species level. As discussed earlier, the human brain architecture, components, sub-systems and processes reflect the underlying environmental factors and feedback loops and emerged from the corresponding mechanisms.

While some of the basic neural system organizations and mechanisms evolved more than once and are shared across species, there is no universality in terms of the overall architecture. As an example, human brain architecture is fundamentally different from the more distributed octopus brain, each shaped by the evolutionary forces, environmental factors and feedback loops [234]. Further, studies have shown that environmental factors and habitat differences cause subtle architectural diversity within the brains of different groups within the same species [235].

Though the human brain is well adapted to its environment, it does not have universally superior capabilities compared to other animals (e.g. acoustic image communication in dolphins, or the superior working memory of numerals in chimpanzees [233]). It encodes a large number of environmental factors and feedback loops in its architecture, though it is seen as a template for the universal AGI. For example, several survival mechanisms pose significant risks from the perspective of AI alignment.

### 7.1. *Universal AGI vs. Heterogeneous AGIs*

Systematic AGI, SAGI, uses these observations to steer away from the pursuit of the "Universal AGI Architecture". It highlights that the universal AGI architecture goals should be replaced with the systematic approach to reach a range of AGIs for different applications. It proposes *AGI Heterogeneity* similar to the natural examples.

Systematic AGI approach tries to capture both (i) shared mechanisms and architectural components and (ii) architectural customization through the shared system design process. As an example, a resource constrained robotics AGI architecture will be different from a large-scale and client-facing AGI on the cloud. Similarly, the actuation, sensor processing subsystems, and the corresponding processes in the robotic setting will be different from the commercial AGI use case. This is due to the differences in the application goals, design constraints, resource constraints as well as the feedback loops (as in multitude of reinforcement learning processes used for training).

### 7.2. *Refocusing on System Design and Feedback Mechanisms*

Systematic AGI focuses on the system design criteria such as system goals, environmental settings, feedback loops, resource constraints and others to determine the architecture, system components, architectural blocks, sub-system designs, learning mechanisms, arbitration controls, and other artificial neural processes for individual use cases.

From this perspective, even the "symbolic", "hybrid" AGI approaches may be replaced with systematic AGI, such that the level and depth of symbolicism, the type and diversity of symbolic mechanisms, the composition and other characteristics as well as their interactions can be highly complex and system dependent.

*It asserts that the system design process is the primary defining factor for AGI, where the symbolicisms and hybridity are emergent characteristics and do not define the AGI.* The architecture of AGI is also a byproduct of the system design mechanisms such as feedback loops instead of being the ultimate goal reached through other mechanisms.

### 7.3. *Heterogeneity in AGI Architecture & Homogeneity in Process*

Though systematic AGI proposes heterogeneity at the architecture level, it proposes homogeneity in the system design processes, goals, regulatory, safety and alignment mechanisms. Despite the diversity of architectures, each AGI architecture training should be highly focused on alignment, safety, compliance through standardized training and feedback loops, protocols and design constraints. The system design processes such as RL loops exert ethics and alignment goals.

### 7.4. *Requirements for Systematic AGI: Learning System Architecture Through Training*

The fundamental difference between this approach and the current AI approaches is that systematic AGI training shapes the entire system architecture, resources, capabilities, individual building blocks, subsystems and processes.

The proposed approach relies on the process such as design goals, constraints, multitude of reinforcement learning loops and processes to reach the target AGI architecture. The level and diversity of symbolic capabilities, system components, sub-systems and interactions emerge from the processes. Hence, should not be set by humans. The process requires a number of feedback loops for alignment, safety, ethics and others in addition to application specific goals and feedback loops.

### 7.5. *Benefits of Substrate & Hardware Utilization for Systematic AGI*

The implementation of systematic AGI requires flexibility in artificial neural networks, such that the system can learn not only the individual neuron parameters, but the architecture and the organization, each of which can be learned and updated through feedback mechanisms, a collection of training lengths in a continual fashion. This is in contrast to the current AI models, in which the system architecture, as in mixture-of-experts (MoE) models, many hyper-parameters and neural network sizes for individual subsystems are fixed and set by humans. Only some of the neural parameters such as weights are learned through gradient descent.

The current separation of substrate from AI capabilities is a limiting factor in contemporary AI architectures which use general purpose servers, GPUs and TPUs to run AI as a "Software-only Solution". Direct hardware training with the goal of learning the entire AGI system (the entire stack from the system and architecture to the functional components) from the design specifications, goals and processes is likely to provide benefits. This requires a paradigm shift in AI towards focusing more on the system design processes than the products that emerge out of the processes.

## 8. Conclusions

Today's AI solutions face a trifecta of challenges including unprecedented levels of energy consumption, safety and ethics problems and roadblocks in reaching artificial general intelligence from narrow AI. The lack of system design contributes to all three grand challenges. This paper argues that artificial intelligence can be reached through multiple design specific paths instead of a universal AGI architecture.

In fact, AGI systems may have diverse architectures and capabilities depending on the use cases. It argues that the focus should be on using system design principles instead of focusing on AGI architectures.

We present Systematic AI for AGI, SAGI, which relies on system design principles and processes to enable a wide range of AGI solutions. Systematic AGI provides novel solution approaches to today's energy efficiency challenges, as well as the leap to aligned AGI. We argue that even the level of symbolicist and connectionist are emergent characteristics from the learning substrate and self-learning system architecture.

AGI requires system design principles to be applied pervasively across all design stages and all aspects of the design. It is required to integrate a number of functional subsystems to operate in a balanced manner, regardless of whether these capabilities are human-like or not.

Alignment, widely acknowledged as the most significant challenge in AI, is the aspect most reliant on system architecture and acts as the principal catalyst underpinning AGI as a foundational requirement. Effectively addressing alignment necessitates architectural frameworks capable of modeling the nuanced complexities of moral reasoning and the extensive ethical processing occurring at all levels, achieving a degree of reliability that surpasses human moral decision-making. We maintain that AGI development is fundamentally contingent upon a learning substrate—one that can adapt its own architecture and specialized functions—and on system design methodologies such as adaptive feedback mechanisms, which go beyond traditional symbolic and emergentist paradigms. Consequently, the AGI produced is not an accidental emergent phenomenon but the outcome of intentional, systematic design and engineering, featuring essential components like an embedded moral architecture integrated deeply within the overall system design.